\documentclass[runningheads]{llncs}
\usepackage{graphicx}
\usepackage{amssymb}
\usepackage{amsmath}
\newcommand{\myvector}[1]{\mathbf{#1}}
\newcommand{\set}[1]{\mathcal{#1}}
\newcommand{\realdim}[1]{\mathbb{#1}}
\usepackage{multirow}
\usepackage[table,xcdraw]{xcolor}

\usepackage{hyperref}
\begin{document}
\title{Mesh2SSM: From Surface Meshes to Statistical Shape Models of Anatomy}
\titlerunning{Mesh2SSM}
%
\author{Krithika Iyer\inst{1,2} \and
Shireen Elhabian\inst{1,2}}
%
\authorrunning{K. Iyer and S. Elhabian.}
%
\institute{Scientific Computing and Imaging Institute, University of Utah, SLC, UT, US \and
Kahlert School of Computing, University of Utah, Salt Lake City, UT, USA
\email{krithika.iyer@utah.edu } \email{shireen@sci.utah.edu}}
\maketitle              
\begin{abstract}
Statistical shape modeling is the computational process of discovering significant shape parameters from segmented anatomies captured by medical images (such as MRI and CT scans), which can fully describe subject-specific anatomy in the context of a population. The presence of substantial non-linear variability in human anatomy often makes the traditional shape modeling process challenging. Deep learning techniques can learn complex non-linear representations of shapes and generate statistical shape models that are more faithful to the underlying population-level variability. However, existing deep learning models still have limitations and require established/optimized shape models for training. We propose Mesh2SSM, a new approach that leverages unsupervised, permutation-invariant representation learning to estimate how to deform a template point cloud to subject-specific meshes, forming a correspondence-based shape model. Mesh2SSM can also learn a population-specific template, reducing any bias due to template selection. The proposed method operates directly on meshes and is computationally efficient, making it an attractive alternative to traditional and deep learning-based SSM approaches.

\end{abstract}
\section{Introduction}

Statistical shape modeling (SSM) is a powerful tool in medical image analysis and computational anatomy to quantify and study the variability of anatomical structures within populations. SSM has shown great promise in medical research, particularly in diagnosis \cite{khan2022machine,schaufelberger2022radiation}, pathology detection \cite{peiffer2022statistical,sophocleous2022feasibility}, and treatment planning \cite{vicory2022statistical}. SSM has enabled researchers to better understand the underlying biological processes, leading to the development of more accurate and personalized diagnostic and treatment plans \cite{merle2019high,bruse2016statistical,lindberg2012hippocampal,faghih20134d}. 

Over the years, several SSM approaches have been developed that implicitly represent the shapes (deformation fields \cite{durrleman2014morphometry}, level set methods \cite{samson2000level}) or explicitly represent them as a ordered set of landmarks or \textit{correspondence points} (aka point distribution models, PDMs). 
Here, we focus on the automated construction of PDMs because, compared to deformation fields, point correspondences are easier to interpret by clinicians,  are computationally efficient for large datasets, and less sensitive to noise and outliers than deformation fields \cite{cerrolaza2019computational}.

SSM performance depends on the underlying process used to generate shape correspondences and the quality of the input data. Various correspondence generation methods exist, including non-optimized landmark estimation and parametric and non-parametric correspondence optimization. Non-optimized methods manually label a reference shape and warp the annotated landmarks using registration techniques  \cite{paulsen2002building,heitz2005statistical,mcinerney1996deformable}. Parametric methods use fixed geometrical bases to establish correspondences \cite{styner2006framework}, while group-wise non-parametric approaches find correspondences by considering the variability of the entire cohort during the optimization process. Examples of non-parametric methods include particle-based optimization \cite{cates2017shapeworks} and Minimum Description Length (MDL) \cite{davies2002learning}.

Traditional SSM methods assume that population variability follows a Gaussian distribution, which implies that a linear combination of training shapes can express unseen shapes. However, anatomical variability can be far more complex than this linear approximation, in which case nonlinear variations normally exist (e.g., bending fingers, soft tissue deformations, and vertebrae with different types). 
Furthermore, conventional SSM pipelines are computationally intensive, where inferring PDMs on new samples entail an optimization process.
Deep learning-based approaches for SSM have emerged as a promising avenue to overcoming these limitations. Deep learning models can learn complex non-linear representations of the shapes, which can be used to generate shape models.  
Moreover, they can efficiently perform inference on new samples without computation overhead or re-optimization. 
Recent works such as FlowSSM \cite{ludke2022landmark}, ShapeFlow \cite{jiang2020shapeflow}, DeepSSM \cite{bhalodia2018deepssm}, and VIB-DeepSSM \cite{adams2022images}  have incorporated deep learning to generate shape models. FlowSSM \cite{ludke2022landmark} and ShapeFlow \cite{jiang2020shapeflow} operate on surface meshes and use neural networks to parameterize the deformations field between two shapes in a low dimensional latent space and rely on an encoder-free setup. Encoder-free methods randomly initialize the latent representations for each sample that are then optimized to produce the optimal deformations. One major caveat of an encoder-free setup is that inference on new meshes is no longer straightforward; the latent representation has to be re-optimized for every new sample. On the other hand, DeepSSM \cite{bhalodia2018deepssm}, TL-DeepSSM \cite{bhalodia2018deepssm}, and VIB-DeepSSM \cite{adams2022images} learn the PDM directly from unsegmented CT/MRI images, and hence alleviate the need for PDM optimization given new samples and can bypass anatomy segmentation by operating directly on unsegmented images. However, these methods rely on supervised losses and require volumetric images, segmented images, and established/optimized PDMs for training. This reliance on supervised losses introduces linearity assumptions in generating ground truth PDMs. TL-DeepSSM \cite{bhalodia2018deepssm}, a variant of DeepSSM \cite{bhalodia2018deepssm}, differs from the others by not utilizing PCA scores as shape descriptors. Instead, it adopts an established correspondence model hence, similar to the vanilla DeepSSM \cite{bhalodia2018deepssm} learns a linear model.

In this paper, we introduce Mesh2SSM \footnote{Source code: \href{https://github.com/iyerkrithika21/mesh2SSM_2023}{https://github.com/iyerkrithika21/mesh2SSM\_2023}}, a deep learning method that addresses the limitations of traditional and deep learning-based SSM approaches. Mesh2SSM leverages  \textit{unsupervised, permutation-invariant representation learning} to learn the low dimensional nonlinear shape descriptor directly from mesh data and uses the learned features to generate a correspondence model of the population. Mesh2SSM also includes an analysis network that operates on the learned correspondences to obtain a data-driven template point cloud (i.e., template point cloud), which can replace the initial template, and hence reducing the bias that could arise from template selection. Furthermore, the learned representation of meshes can be used for predicting related quantities that rely on shape. Our main contributions are:
\begin{enumerate}
    \item We introduce Mesh2SSM, a fully unsupervised correspondence generation deep learning framework that operates directly on meshes. Mesh2SSM uses an autoencoder to extract the shape descriptor of the mesh and uses this descriptor to transform a template point cloud using IM-Net \cite{chen2019net}. 
    \item The proposed method uses an autoencoder that combines geodesic distance features and EdgeConv \cite{wang2019dynamic} (dynamic graph convolution neural network) to extract meaningful feature representation of each mesh that is permutation-invariant. 
    \item Mesh2SSM also includes a variational autoencoder (VAE)  \cite{kingma2013auto,rezende2014stochastic} operating on the learned correspondence points and trained end-to-end with correspondence generation network. This VAE branch serves two purposes: (a) serves as a shape analysis module for the non-linear shape variations and (b) learns a data-specific template from the latent space of the correspondences that is fed back to the correspondence generation network. 
\end{enumerate}

\begin{figure}
    \includegraphics{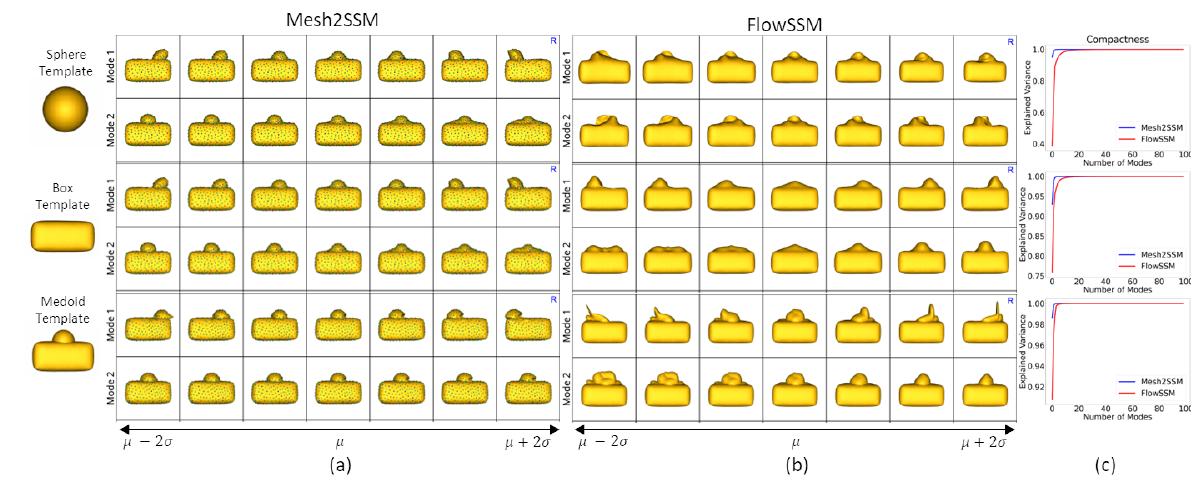}
    \caption{Top two PCA modes of variations identified by (a) Mesh2SSM and (b) FlowSSM \cite{ludke2022landmark} with three templates: sphere, box without a bump, and medoid shape. FlowSSM fails to capture the horizontal movement as the primary mode of variation. (c) The compactness curves for both models with different templates.}
    \label{fig:teaser}
    
\end{figure}

To motivate the need for the mesh feature encoder and study the effect of the template selection, we considered the box-bump dataset, a synthetic dataset of 3D shapes of boxes with a moving bump. In Figure~\ref{fig:teaser}, we compare Mesh2SSM (sans the VAE analysis branch) with FlowSMM \cite{ludke2022landmark} since this approach is the closest to Mesh2SSM. We performed experiments with three templates: medoid, sphere, and box without the bump. Although both methods show some sensitivity to the choice of template, FlowSSM is more sensitive toward the choice of the template than Mesh2SSM. Moreover, FlowSSM fails to identify the correct mode of variation, the horizontal movement of the bump as the primary variation, which can also be inferred by comparing the compactness curves in Figure~\ref{fig:teaser}.c. Mesh2SSM performs best when the template is a medoid shape, which makes the case for learning a data-specific template. Since Mesh2SSM model uses an autoencoder, inference on unseen meshes only requires a single forward pass (1 second per sample); FlowSSM requires re-optimization, increasing the inference time drastically and require a convergence criteria to determine the best number of iterations per sample (0.15 seconds for one iterations per sample). 

\section{Method}
\begin{figure}[!ht]
    \centering
    \includegraphics[scale=0.72]{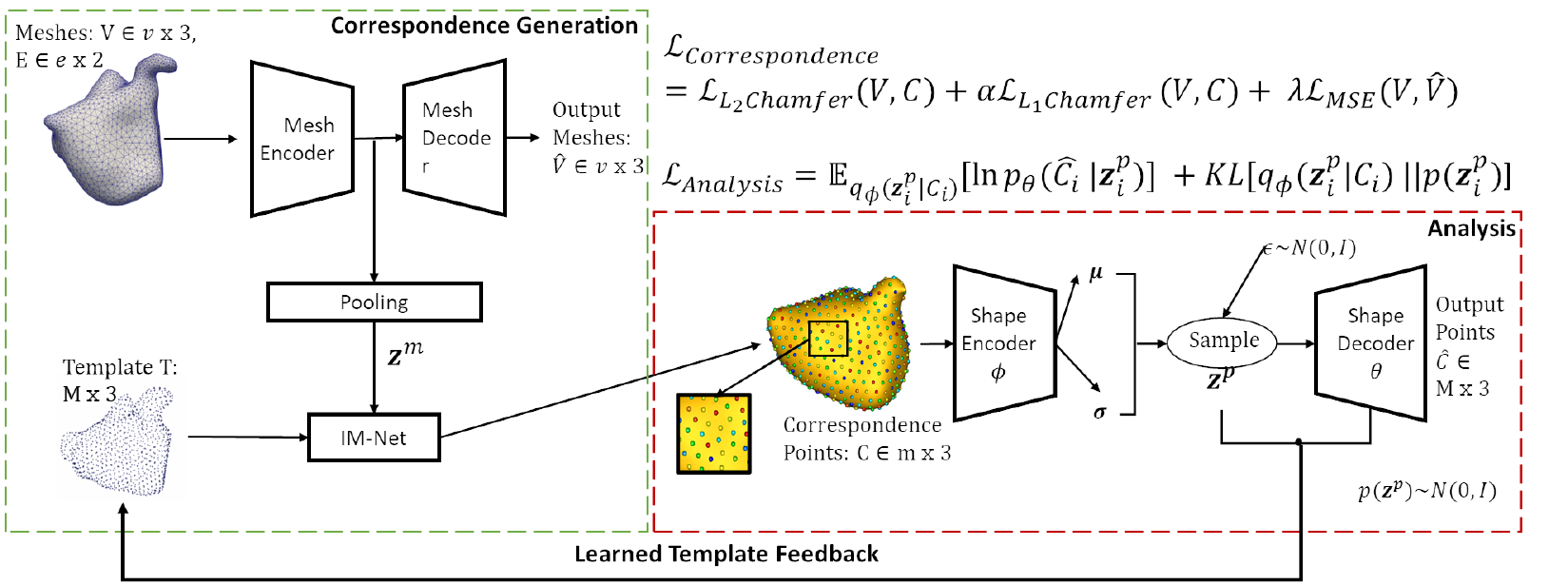}
    \caption{\textbf{Mesh2SSM:} Architecture and loss of the proposed method.}
    \label{fig:mesh2ssm_model}
\end{figure}

The overview of the proposed pipeline is provided in Figure~\ref{fig:mesh2ssm_model}. This section provides a brief description of each module.

\subsection{Correspondence Generation}
Given a set of \(N\) aligned surface meshes \(\set{X} = \{X_1,X_2,...X_N\}\), each mesh \(X_i = (V_i, E_i)\), where \(V_i\) and \(E_i\) represent the vertices and edge connectivity, respectively. The goal of the model is to predict a set \(C_i\) of \(M\) \(3D\) correspondence points that fully describe each surface \(X_i\) and are anatomically consistent across all meshes. This goal is achieved by learning a low dimensional representation of the surface mesh \(\myvector{z}^m \in \realdim{R}^L\) using the mesh autoencoder and then \(\myvector{z}^m\) is used to transform the template point cloud via the implicit field decoder (IM-Net) \cite{chen2019net}. The network optimization is driven primarily by point-set to point-set two-way Chamfer distance between the learned correspondence point sets \(C_i\) and the vertex locations \(V_i\) of the original meshes. To ensure that the encoder learns useful features for the task, we regularize the optimization using the vertex reconstruction loss of the autoencoder between the input \(V_i\) and the predicted \(\hat{V_i}\). The correspondence loss function is given by:
\begin{equation}\label{mesh2ssm_equation_1}
    \mathcal{L}_{C} = \sum_{i=1}^N \left[\mathcal{L}_{L_2 Chamfer}(V_i,C_i) + \alpha \mathcal{L}_{L_1 Chamfer}(V_i,C_i)  + \gamma \mathcal{L}_{MSE}(V_i,\hat{V_i}) \right]
\end{equation}
where \(\alpha,\gamma\) are the hyperparameters. We consider a combination of \(L_1\) and \(L_2\) two-way Chamfer distance for numerical stability as the magnitude of \(L_2\) loss can be low over epochs and \(L_1\) can compensate for it. The correspondence generation uses two networks:\\
\textbf{Mesh Autoencoder (M-AE):} We use EdgeConv \cite{wang2019dynamic} blocks, which are dynamic graph convolution neural network (DGCNN) blocks in the encoder and decoder to capture local geometric features of the mesh. The model takes vertices as input, computes an edge feature set of size \(k\) (using nearest neighbors) for each vertex at an EdgeConv layer, and aggregates features within each set to compute EdgeConv responses. The output features of the last EdgeConv layer are then globally aggregated to form a 1D global descriptor \(\myvector{z}^m_i\) of the mesh. The first EdgeConv block uses geodesic distance on the surface of the mesh to calculate the \(k\) features. The dynamic feature creation property of EdgeConv and the global pooling make this autoencoder permutation invariant.\\ %
\textbf{Implicit field decoder (IM-NET):} The IM-NET \cite{chen2019net} architecture consists of fully connected layers with non-linearity and skip-layer connections. This network enforces the notion of correspondence across the samples. The network takes in two inputs, the latent representation of the mesh \(\myvector{z}^m\) and a template point cloud (a set of unordered points). IM-NET estimates the deformation of each point in the template required to deform the template to each sample, conditioned on \(\myvector{z}^m\). Based on the learned deformation, IM-NET directly produces the resultant displaced template point without the computational complexity of the deformation fields. Correspondence is established since the same template is deformed to all the samples.

\subsection{Analysis}
The Mesh2SSM model also consists of an analysis branch that acts as a shape analysis module to capture non-linear shape variations identified by the learned correspondences \(\{C_i\}_{i=1}^{N}\) and also learns a data-informed template from the latent space of correspondences to be fed back into the correspondence generation network during training. This branch uses one network module:\\
\textbf{Shape Variation Autoencoder (SP-VAE):} The VAE \cite{kingma2013auto,rezende2014stochastic} is a latent variable model parameterized by an encoder \(\phi\), decoder \(\theta\), and the prior \(p(\myvector{z}^p) \sim \mathcal{N}(0,\mathbf{I})\). The encoder maps the shape represented by the learned correspondence points \(C\) to the latent space and the decoder reconstructs the correspondences from the latent representation \(\myvector{z}^p\). By capturing the underlying structure of the PDM through a low-dimensional representation, SP-VAE allows for the estimation of the mean shape of the learned correspondences. The SP-VAE is trained using the loss function given by:
    \begin{equation}\label{vae_equation}
    \mathcal{L(\theta,\phi)} = -\mathbb{E}_{q_{\phi} (\myvector{z}^p_i|C_i)} \left[\operatorname{log} p_\theta(C_i|\myvector{z}^p_i)\right] + KL(q_{\phi}(\myvector{z}^p_i|C_i) || p(\myvector{z}^p_i))      
    \end{equation}

The main difference between M-AE and a SP-VAE lies in the input and output representations they handle.
SP-VAE operates directly on sets of landmarks or correspondences, aiding in the analysis of shape models. It takes a set of correspondences describing a shape as input and aims to learn a compressed latent representation of the shape. Importantly, the SP-VAE maintains the same ordering of correspondences at the input and output, so it does not use permutation-invariant layers or operations like pooling.
.

\subsection{Training}
We begin with a burn-in stage, where only the correspondence generation module is trained while the analysis module is frozen. After the burn-in stage, alternate optimization of the correspondence and analysis module begins. 
During the alternate optimization phase, we generate the data-informed template from the latent space of SP-VAE at regular intervals. The learned data-informed template is used in the correspondence generation module in the subsequent epochs. For the learned template, we sample 500 samples from the prior \(p(\myvector{z}^p) \sim \mathcal{N}(0,\mathbf{I})\) and pass it through the decoder of SP-VAE to get the reconstructed correspondence point set. The mean template is defined by taking the average of these generated samples. Inference with unseen meshes is straight forward; the meshes are passed through the mesh encoder and IM-NET of the correspondence generation module to get the predicted correspondences. All hyperparameters and network architecture details are mentioned in the supplementary material. 
\section{Experiments and Discussion}

\textbf{Dataset:} We use the publicly available Decath-Pancreas dataset of 273 segmentations from patients who underwent pancreatic mass resection \cite{simpson2019medseg}. The shapes of the pancreas are highly variable and have thin structures, making it a good candidate for non-linear SSM analysis. The segmentations were isotropically resampled, smoothed, centered, and converted to meshes with roughly 2000 vertices. Although the DGCNN mesh autoencoder used in Mesh2SSM does not require the same number of vertices, uniformity across the dataset makes it computationally efficient; hence, we pad the smallest mesh by randomly repeating the vertices (akin to padding image for convolutions). The samples were randomly divided, with 218 used for training, 26 for validation, and 27 for testing.

\subsection{Results}
We perform experiments with two templates: sphere and medoid. We compare the performance of FlowSSM \cite{ludke2022landmark} with Mesh2SSM with the template feedback loop. For Mesh2SSM template, we use 256 points uniformly spread across the surface of the sample. Mesh2SSM and FlowSSM do not have a equivalent latent space for comparison of the shape models, hence, we consider the deformed mesh vertices of FlowSSM as correspondences and perform PCA analysis.  
\begin{figure}[t]
    \centering
    \includegraphics[scale=0.95]{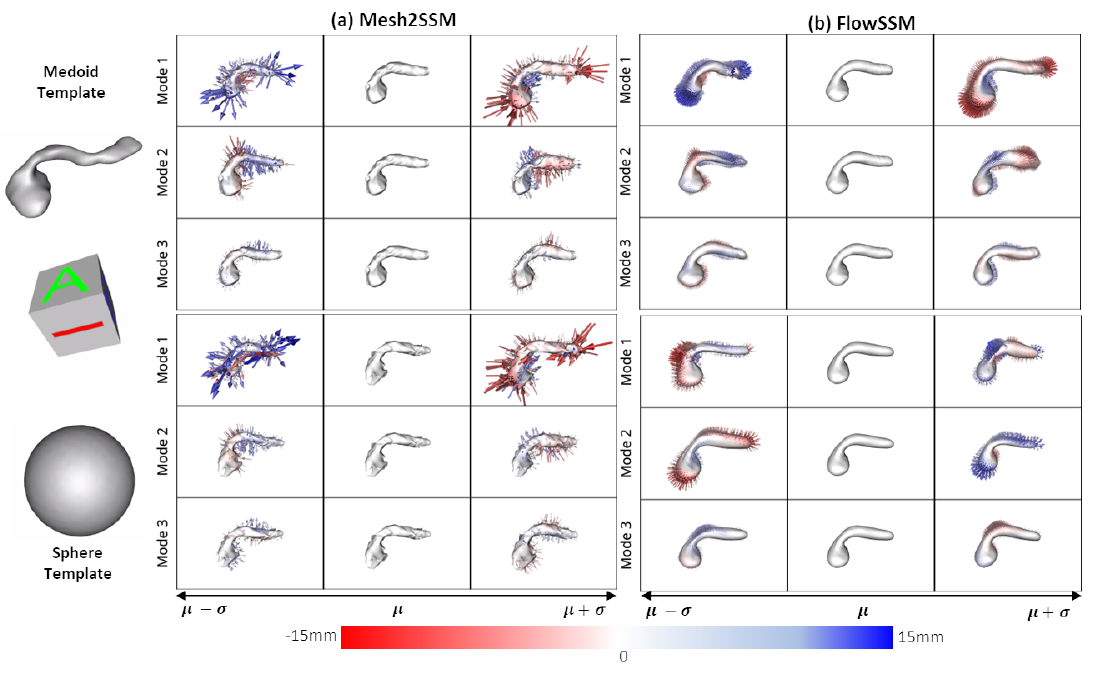}
    \caption{Top three PCA modes of variations identified by (a) Mesh2SSM and (b) FlowSSM \cite{ludke2022landmark} with two templates: sphere, medoid. The color map and arrows show the signed distance and direction from the mean shape.}
    \label{fig:pancreas_modes}
\end{figure}
Figure~\ref{fig:pancreas_modes} shows the top three PCA modes of variations identified by Mesh2SSM and FlowSSM. Similar to the observations made box-bump dataset, FlowSSM is affected by the choice of the template, and the modes of variation differ as the template changes. On the other hand, PDM predicted by Mesh2SSM identifies the same primary modes consistently. Pancreatic cancer mainly presents itself on the head of the structure \cite{ralston2018davidson} and for the Decath dataset, we can see the first mode identifies the change in the shape of the head. 
\begin{figure}[t]
    \centering
    \includegraphics[scale=0.8]{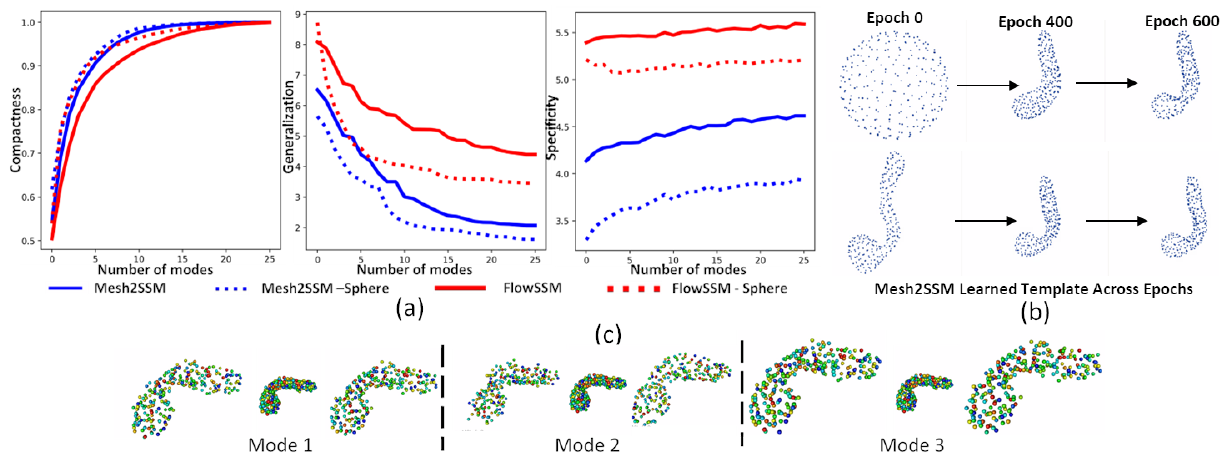}
    \caption{(a) Shape statistics of pancreas dataset: compactness (higher is better), generalization (lower is better), and specificity (lower is better). (b) Mesh2SSM Learned template across epochs for pancreas dataset. (c) Non-linear modes of variations identified by Mesh2SSM. }
    \label{fig:pancreas_stats}
\end{figure}
We evaluate the models based on compactness, generalization, and specificity. Compactness measures the ability of the model to reconstruct new shape instances with fewer parameters using PCA explained variance. Generalization measures the average surface distance between all test shapes and their reconstructions, and specificity measures the distance between randomly generated PCA samples. Figure~\ref{fig:pancreas_stats}.a shows the metrics for the pancreas dataset. Mesh2SSM outperforms FlowSSM in all three metrics, despite using only 256 correspondence points compared to FlowSSM's \(\sim\)2000 vertices. Mesh2SSM correspondence generation module efficiently parameterizes the surface of the pancreas with a minimum number of parameters. Mesh2SSM template, shown in Figure~\ref{fig:pancreas_stats}.b, becomes more detailed as optimization continues, regardless of the starting template. The model can learn correct deformations in the correspondence generation module and identify the correct mean shape in the latent space of SP-VAE in the analysis module. Using the analysis module of Mesh2SSM, we visualized the top three modes of variation identified by by sorting the latent dimensions of SP-VAE based on the standard deviations of the latent embeddings of the training dataset. Variations are generated by perturbing the latent representation of a sample in three directions, resulting in non-linear modes such as changes in the size and shape of the pancreas head and narrowing of the neck and body. This is shown in Figure~\ref{fig:pancreas_stats}.c for MeshSSM model with medoid starting template. 
\begin{table}[!t]
    \centering
    \caption{Distance metrics (measured in mm) of the testing samples and their reconstructions for the pancreas dataset}
    \resizebox{0.9\textwidth}{!}{
    \begin{tabular}{|c|c|c|c|c|}
    \hline
        ~ & \multicolumn{2}{c|}{Mesh2SSM}  & \multicolumn{2}{c|}{FlowSSM \cite{ludke2022landmark}}  \\ \hline
         \multirow{2}{*}{Metrics} ~ & \multicolumn{4}{c|}{Template}   \\ \cline{2-5}
       ~& Medoid & Sphere & Medoid & Sphere \\ \hline
        \(L_1\) Chamfer  & \textbf{0.033 \(\pm\) 0.002} & 0.035 \(\pm\) 0.002 & 0.391 \(\pm\) 0.162 & 1.91 \(\pm\) 0.687 \\ 
        Surface-to-Surface  & \textbf{2.378 \(\pm\) 0.7325} & 5.436 \(\pm\) 2.232 & 5.918 \(\pm\) 2.026 & 4.918 \(\pm\) 1.925 \\ \hline
    \end{tabular}} 
    \label{distance_table}
\end{table}
The distance metrics for the reconstructions of the testing samples were also computed. The results of the metrics are summarized in Table~\ref{distance_table}. The calculation involved the \(L_1\) Chamfer loss between the predicted points (correspondences in the case of Mesh2SSM and the deformed mesh vertices in the case of FlowSSM) and the original mesh vertices. Additionally, the surface to surface distance of the mesh reconstructions (using the correspondences in Mesh2SSM and deformed meshes in FlowSSM) was included. For the pancreas dataset with the medoid as the initial template, Mesh2SSM with the template feedback produced more precise models.
\subsection{Limitations and Future Scope}
As SSM is included a part of diagnostic clinical support systems, it is crucial to address the drawbacks of the models. Like most deep learning models, performance of Mesh2SSM could be affected by small dataset size, and it can produce overconfident estimates. An augmentation scheme and a layer uncertainty calibration are could improve its usability in medical scenarios. Additionally, enforcing disentanglement in the latent space of SP-VAE can make the analysis module interpretable and allow for effective non-linear shape analysis by clinicians.

\section{Conclusion}
The paper presents a new systematic approach of generating non-linear statistical shape models using deep learning directly from meshes, which overcomes the limitations of traditional SSM and current deep learning approaches. The use of an autoencoder for meaningful feature extraction of meshes to learn the PDM provides a versatile and scalable framework for SSM. Incorporating template feedback loop via VAE  \cite{kingma2013auto,rezende2014stochastic} analysis module helps in mitigating bias and capturing non-linear characteristics of the data. The method is demonstrated to have superior performance in identifying shape variations using fewer parameters on synthetic and clinical datasets. To conclude, our method of generating highly accurate and detailed models of complex anatomical structures with reduced computational complexity has the potential to establish statistical shape modeling from non-invasive imaging as a powerful diagnostic tool.

\subsubsection{Acknowledgements}
This work was supported by the National Institutes of Health under grant numbers NIBIB-U24EB029011, NIAMS-R01AR076120, and NHLBI-R01HL135568. We thank the University of Utah Division of Cardiovascular Medicine for providing left atrium MRI scans and segmentations from the Atrial Fibrillation projects and the ShapeWorks team.

\clearpage

\bibliographystyle{splncs04.bst}
\bibliography{paper3629_ref}
\section{Supplementary}
\subsection{Architecture}

\begin{figure}
        \centering
        \includegraphics[scale=1.25]{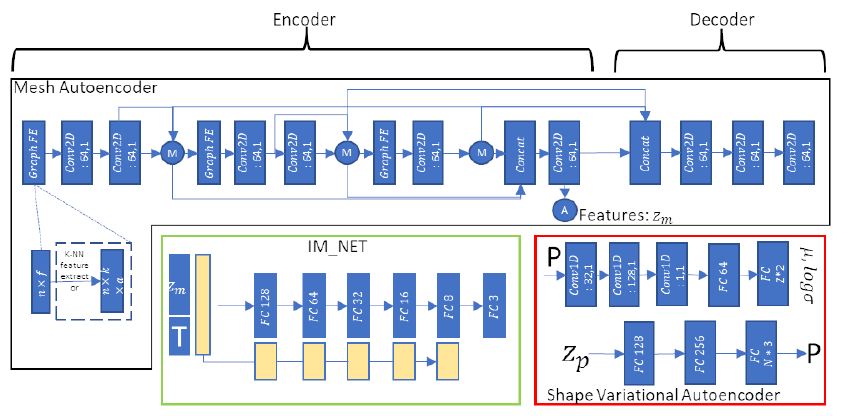}
    \caption{Architecture for (a) mesh autoencoder where each convolutional block is followed by non-linear activation and batch normalization, (b) IM-NET where each fully connected layer is followed  by activation function, and (c) shape variational autoencoder. M: max-pool, A: average-pool, T: each point from the template point cloud.}
    \label{fig:mesh2smm_arch}
\end{figure}

\begin{enumerate}
    \item FlowSSM: Used the official implementation provided by the authors at\\ github.com/davecasp/flowssm 
    \item Mesh2SSM:\\
    \textbf{Netowrk}: See Figure~\ref{fig:mesh2smm_arch} for network for (a) mesh autoencoder (MAE), (b) IM-NET, and  (c) point VAE (P-VAE). All networks use leaky RELU. \\
    \textbf{Hyper-parameters:} 
\begin{multicols}{2}
\begin{enumerate}
    \item Learning rate MAE 0.01 with step scheduler
    \item Learning rate P-VAE  0.0009 with step scheduler
    \item Batch size 10
    \item Latent dim \(z_m\) and \(z_p\) = 32 (box-bump), 64 (pancreas), 128 (left atrium)
    \item Epochs: 1000
    \item MAE: \(\alpha = 0.01\) increased gradually to 1, \(\gamma = 0.01\)
    \item GPU: NVIDIA GeForce RTX 2080 Ti
    \item Adam optimizer
\end{enumerate}
\end{multicols}
\end{enumerate}

\subsection{Results: Left Atrium}
Left atrium dataset: 1102 anonymized segmented LGE MRI images from unique atrial fibrillation patients with spatial resolution \(0.65 \times 0.65 \times 2.5 mm^3\). Train, test, validation split: 900/66/136 samples. 

\begin{figure}[!t]
    \centering
    \includegraphics[scale=0.78]{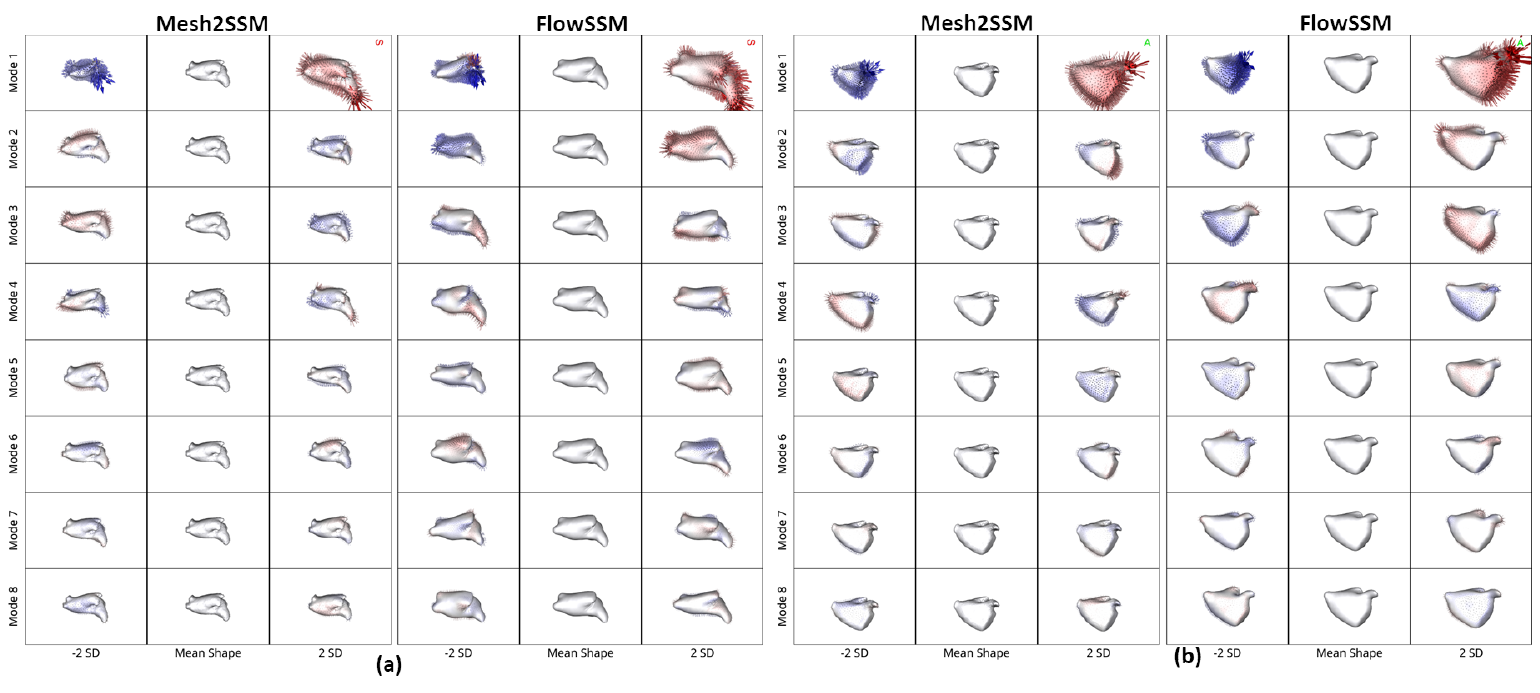}
    \caption{Top eight PCA modes of variations identified by Mesh2SSM and FlowSSM with medoid as template shown from the (a) top view and (b) anterior view. The color map and arrows show the signed distance and direction from the mean shape.}
    \label{fig:my_label}
\end{figure}

\begin{table}[!ht]
    \centering
    \caption{Distance metrics (measured in mm) of the testing samples and their reconstructions for the left atrium dataset with medoid template}
    \resizebox{0.5\textwidth}{!}{
    \begin{tabular}{|c|c|c|}
    \hline
        Metrics & Mesh2SSM  & FlowSSM\\ \hline
        \(L_1\) Chamfer &0.0383 \(\pm\) 0.0026 & 0.2547 \(\pm\) 0.0532\\
        Surface-to-Surface &3.9439 \(\pm\) 0.6997 & 0.2512 \(\pm\) 0.0505\\
        \hline
    \end{tabular}} 
    \label{distance_table_la}
\end{table}

\begin{figure}[!ht]
    \centering
    \includegraphics[scale=0.7]{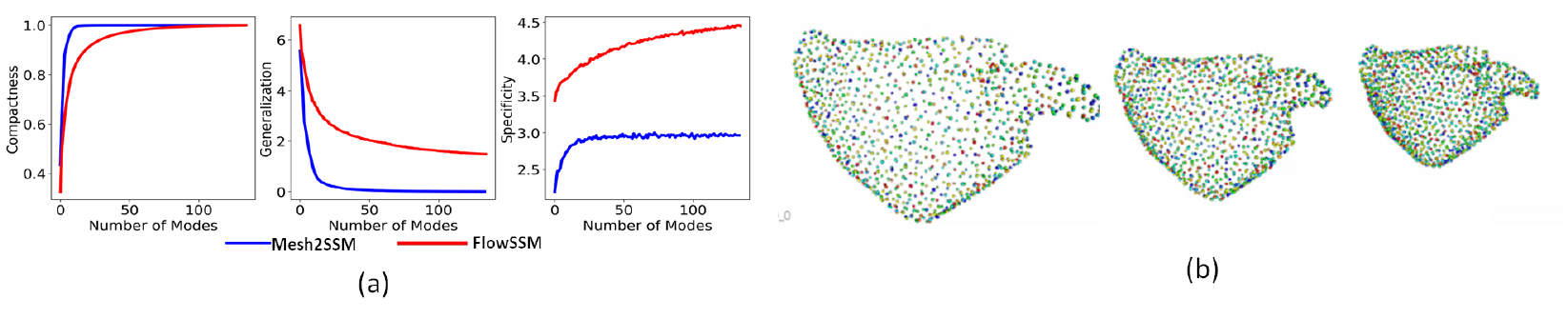}
    \caption{(a) Shape statistics of left atrium dataset: compactness (higher is better), generalization (lower is better), and specificity (lower is better). (b) First dominant non-linear mode of variation identified by Mesh2SSM.}
    \label{fig:la_metrics}
\end{figure}

\end{document}